# About the Unreal

*John Beverley*[a,b,c*], *Jim Logan*[d], *and Barry Smith*[a,b]

[a] *Department of Philosophy, University at Buffalo, Buffalo NY, USA.*
[b] *National Center for Ontological Research, Buffalo NY, USA.*
[c] *Institute for Artificial Intelligence and Data Science, Buffalo NY, USA.*
[d] *Ontogenesis Solutions, LLC*

**Abstract.** We introduce a framework for representing information about entities that do not exist or may never exist, such as those involving fictional entities, blueprints, simulations, and future scenarios. Traditional approaches that introduce "dummy instances" or rely on modal logic are criticized, and a proposal is defended in which the cases in question are modeled using the intersections of actual types rather than specific non-existent tokens. The paper positions itself within the Basic Formal Ontology paradigm and its realist commitments, emphasizing the importance of practical, implementable solutions over purely metaphysical or philosophical proposals. We argue that existing approaches to non-existent entities either overcommit to metaphysical assumptions or introduce computational inefficiencies that hinder applications. By developing a structured ontology-driven approach to unreal patterns, the paper aims to provide a useful and computationally viable means of handling references to hypothetical or non-existent entities.

## 1. Information and Aboutness

Information consists in – at a minimum – copyable patterns that are about something [1–7]. A line of ants that by happenstance form the image of your mother is a copyable pattern. It is not, however, *about* anything [8]. Such a line could, of course, be about something; then, however, you and your cognitive attitudes must arguably be involved [2, 9]. This is to say that there are copyable patterns that are, and copyable patterns that are not, about something. Examples of the former include familiar fare such as coordinate systems, coding paradigms, the content of novels, paintings, poems, and so on. Examples of the latter include brick walls, a single quote mark, a lone universal quantifier, and so on.

While there are threads worth untangling with respect to information thus understood [9], there is a larger Gordian Knot deserving attention. We will focus here on types of copyable patterns that purport to be about something that, in some sense, does not exist or indeed may never exist. Considerable ink has been spilled on this and adjacent topics from the pens of philosophers [10–16], logicians [17–19], and ontologists [20, 21]. Our discussion here falls within the discipline of ontology engineering; more specifically, formal implementations within that field. We thus leave aside concerns over identity conditions, essences, modality, putative possible worlds, and the like, aiming

* Corresponding author.
: johnbeve@buffalo.edu (J. Beverley); jlogan@ontogenesis-solutions.com (Jim Logan); ifomis@gmail.com (Barry Smith)

instead for easily implementable guidance that respects intuitions – both from common sense and from domain experts – regarding such entities. Our proposal is thus to be judged on the basis of whether it provides a formally consistent and practically implementable characterization of the target phenomenon: copyable patterns that purport to be about something that, in some sense, does not exist or indeed may never exist.

## 2. Three Cases

It will be helpful to identify a handful of cases to scope our target, each containing well-trodden examples of information that seems to be about entities of problematic sorts: fictions, blueprints, and simulations.

### *2.1 Fiction*

The erudite and petulant Ignatius Riley [22] from *Confederacy of Dunces* and the elegant Scarlett O'Hara from *Gone with the Wind* [23] are fictional characters. Mythological narratives too often contain reference to fictional characters, such as Homer's depiction of Achilles and Patroclus [24], or fictional locations, such as Midgard in the Eddas [25]. Fiction understood here may also concern real characters, as in the case of *War and Peace*, which is set in Moscow. Or it may treat of hypotheticals such as "What if Nelson had lost to Napoleon at the Battle of Trafalgar", or of counterfactuals such as "If Oswald hadn't killed Kennedy, someone else would have" [26]. There is also speculative technology, such as Ray Bradbury's matryoshka brains [27] or H.G. Wells' anti-gravity cavorite that enabled faster-than-light travel [28]. In each example, we see putative reference to some manner of fictional entity that is assumed to have never existed.

Given this wide range of fictions and of fictional entities, it will be helpful to identify a paradigmatic example on which to focus our attention. Superman, for those unfamiliar, is a fictional character represented in various media. When one hears putatively accurate assertions such as "Superman is strong" there is an implied context in which the expression is to be understood, not to be confused with the real world which we inhabit. Superman is strong in representations within, say, comic issues published by DC Comics. And it is indeed true in the actual world that a Superman character is described in these publications as being strong. To evaluate the claim that "Superman is strong" one must relativize the expression to some seemingly fictional context for evaluation.

### *2.2 Blueprints*

Many digital twins represent physical assets, though some simply serve as prototypes that prescribe how to create a future physical asset. So-called "digital twin instances" mimic physical products to which a representing digital twin remains linked throughout the life of the product, while "digital twin prototypes" specify what must be done to produce a physical product meeting the specifications of the prototype [29]. Similarly, blueprints representing buildings, pharmaceutical chemicals, or vehicles, provide guidance for how certain products can be created, and this is so whether or not they are eventually created. Such guidance need not be focused exclusively on the creation of products, however. Something similar applies also to training protocols, education plans, legal codes, congressional bills, and so on. For example, the *Tax Cuts and Jobs Act of 2017* prescribed a lowering of taxes on corporations and individuals in the US [30] which did not take effect until 2018. Failed legislation seems to sharpen the issue further, as one might plausibly ask what the 1926 *Child Labor Amendment* to the US Constitution is about, given that amendments require 38 states to ratify, and as of 1937 only 28 states had done so [31]. Throughout these examples there is a prescription in play reflecting how someone or some group desires the world to be, whether that involves a change to the way the world is or maintenance regarding how they want the world to continue being.[1]

Our paradigmatic case will center on Samantha, who works for the car manufacturer Honda and is tasked with drafting and promoting specifications for new vehicle models for some coming year. When Samantha creates a blueprint detailing a new Honda Civic model to be produced in 2035, there is an implied prescribed context against the background of which her work should be understood. Samantha is not, we may assume, describing anything that presently exists, but is instead outlining a blueprint for some entity intended to satisfy certain requirements once manufactured; she is prescribing how she intends the world to be. Expressions Samantha makes about the intended output, such as "The Honda Civic SLS 2035 will prove itself to be my best work" are thus evaluated against this special context.

### *2.3 Simulation*

Simulation involves the use of physical, mathematical, behavioral, or logical representations of some system in the interest of simulating how that system may behave under various circumstances [33]. Such representations often concern systems that may or may not exist, such as historical systems, where our knowledge turns on events incompletely recorded, or biological processes which occur at scales too small for direct observation. In other cases, the simulated systems may be entirely theoretical, such as a system devised to study the formation of galaxies under alternative physical constants [34]. Engineers devising simulations for, say, red teaming exercises [35] often do not want the simulated events to occur. Indeed, they are often preparing strategies for preventing the simulated events from coming to pass as well as planning for what steps to take in the event they do. In either case, evaluations of expressions borne out of simulation exercises, such as "Respond with nuclear capabilities" must be evaluated against some special unreal context, as in the case of fiction and blueprints.

As a paradigmatic example, consider a team of cybersecurity experts engaged in red-team exercises regarding potential vulnerabilities in a network. Discourse concerning simulations of this network are to be interpreted within the special context in which certain threats and defense postures may or may not manifest.

### *2.4 Connecting Themes*

There are of course many other similarities and differences across these cases. For example, our everyday cognitive attitudes and conversations involve the uses of expressions which involve engagement of imagination. From another direction, fiction neither describes how the actual world is nor prescribes how one wants the world to be. In contrast, blueprints are often prescriptions for how one wants the world to be, while simulations are often descriptions about how the world could be. Simulations do often involve prescriptions also, as when they are used to inform decisions about what would have to be done should certain conditions obtain.

[1] Compare Searle and Vanderveken on mind-to-world and world-to-mind directions of fit [32].

Expressions about fictional entities are not directed to the future, but blueprints and simulations are. Blueprints are often about desired or desirable outcomes, whereas simulations are often about undesirable outcomes and subsequent planning in the event such outcomes are realized. Simulations are thus characteristically associated with probabilities or likelihoods. While these differences are important, it remains the case that expressions found in all the mentioned cases are connected by relativization to some special context.[2]

Another theme worth highlighting is that expressions stemming from each case are often interpreted in natural language as referencing some instance or some individual. For example, the expression "Superman wears a red cape" suggests reference to an individual called "Superman." There are some who talk as if there were some entity denoted by "Superman", for example they hold that the referent of this expression is a concept, or some other mental particular [36, 37]. But it is not a concept that wears, or is held to wear, a red cape. There is however an aspect of such expressions which does indeed involve reference to entities that exist in the real world, namely to entities existing *at the level of types* rather than of *instances*. Expressions such as "Superman wears a red cape" are about familiar, actual entity types such as **Wearing**[3], and **Redness**, and **Cape**, where an *actual entity type* is a type which is or has been instantiated. And there are numerous examples of all the mentioned types in this the actual world.

These remarks apply just as well to implicit reference in fictional sentences, such as "Superman is from the planet Krypton", where there is reference to types such as **Planet**, **Rocky**, **Watery**, and so on. There is however no reference to any entity denoted by the term "Krypton", because there is no such entity. Similarly, simulations about red teaming exercises involve references to types such as **Bot Network** that have actual instances; an instance of **Blueprint** for a planned Honda Civic SLS 2035 may refer to **Steering Wheel** and **Portion of Metal**, instances of which, again, exist.

We can thus accept that each of our examples (comics, blueprints, and so forth) are *about* something even where they involve no reference to specific instances in the actual world. Indeed, it would be quite odd, we claim, to assert that Samantha creates a blueprint for a Honda Civic SLS 2035 *instance*, prior to any such vehicle being manufactured. Such a commitment opens the door to questions such as "To which instance does the blueprint then refer?"

There is another sense in which the actual world harbors much more that we can faithfully represent than is achievable through fiction, blueprints, or simulations. For in each case the information artifacts in question exhibit *loci of indeterminacy* [16]. There is no answer to the question "What was Superman's grandmother's eye color?" because this topic was never addressed in the Superman narrative. If Superman existed, however, this question would have an answer, and this regardless of whether it is or is not known by anyone. As diligent as Samantha is in construction of a blueprint for her new model, there will be dimensions of variability between what is specified in her plan and any actual vehicles produced. Red- and blue-teaming simulations, similarly, cannot hope to characterize all unknown unknowns. The contexts in relation to which they are defined are necessarily incompletely specified.

---

[2] Most, if not all, expressions are evaluated according to *some* context, e.g. "The refrigerator is empty" is relativized to my refrigerator on the one hand and to some amorphous level of granularity on the other. It would be inappropriate for one to respond, "There are molecules in your refrigerator."

[3] We adopt the convention of displaying ontology classes capitalized in bold, relations are italicized, and instances underlined.

### 3. OWL and Dummy Instances

How, then, are we to formally interpret statements under each of our three headings to extract the relevant knowledge which they contain? How, more precisely, are we to do this using the direct semantic version of the Web Ontology Language (OWL2DL). OWL2DL extends the Resource Description Framework (RDF) and RDF Schema (RDFs), which represent data as sets of subject-predicate-object directed graphs. OWL2DL supplements these languages by providing a logical vocabulary allowing for expression of relationships, such as when two classes are disjoint. In every case, such logical vocabulary is interpreted in terms of relationships among instances which fall under classes. For example, to say that class A is disjoint from class B is to say these classes share no instances. OWL2DL reflects a decidable fragment of first-order logic, meaning there is an algorithm which can determine the truth-value for any statement expressed in the language in a finite number of steps, a feature that is important for ontology evaluation using automated reasoners such as HermiT [38].

Asserting relationships in OWL2DL is achieved using object properties (aka relations), which hold between instances. OWL2DL adopts the standard interpretation of universal and existential quantifiers, insofar as the former range over any instance in the domain and the latter range over some instance in the domain. For example, we would paraphrase "All humans are mammals" along the lines of "Any instance of the class **Human** is an instance of the class **Mammal**." Similarly, we would paraphrase "All organisms have some cell as part" along the lines of "Any instance of the class **Organism** stands in a *has part* relationship to some instance of the class **Cell**."

Some authors, leveraging OWL2DL and accepting that information must in every case be *about* something, maintain – in contrast to the discussion above – that we must introduce instances to adequately characterize the knowledge conveyed by expressions in fiction, blueprints, simulations. This is not to say these authors believe that expressions such as "Superman wears a red cape" refer to some actual instance of Superman in the world, something over and beyond a mental representation or literary description. Rather, these authors are motivated to introduce such "dummy" instances to facilitate reasoning or to ease information extraction, owing to the OWL2DL restriction that object properties must relate only instances. For example, a straightforward way to represent that Superman lives in Metropolis is to leverage an object property such as *is an inhabitant of*; this would require, however, that a Superman instance relates to a Metropolis instance.

Various strategies have emerged to characterize such relationships. Some urge that the so-called "punning" strategy should be employed to create a link between a class name and an instance that falls under an object property for the purpose of consistent reasoning [39]. However, strategies of this sort do not relate classes and instances in the context of reasoning, as is well known, and so would require additional technical and theoretical support to be made feasible. Similarly, some suggest that, for example, an instance of a blueprint is about some entity that is of the same type as instances *would be* were they to exist [40]. The most developed version of this proposal draws a distinction between planned and actual instances, where the former is what is prescribed by a blueprint and the latter what would (ultimately) be created according to the prescription. Planned and actual instances are connected by a 'counterpart' relation and asserted to fall under the same parent class. Such a proposal leads to confusion, however, as to talk of any actual instance is to talk of something in the real world, where when we talk of a planned instance there is nothing of this sort that is denoted. Consider, too, that while an actual Honda Civic instance is presumably an instance of the class **Material Entity**, since it has matter as parts, no planned instance has material parts, and so no planned instance should be considered an instance of **Material Entity**; yet the mentioned proposal entails such a classification. It is, lastly, unclear how the proposal generalizes to our other cases.

There remain other strategies. For example, some urge that blueprints are about some specific instances though they are carved off from the actual world into some "modal" context that is otherwise quite like the actual world [41]. Strategies of this sort seem generalizable to fiction as well as simulation. But they conflict with the intuition discussed above to the effect that Samantha does not create a blueprint for any specific instance of **Vehicle** to be created that would go beyond the realm of mental representation or description. What goes for blueprints seems to hold for the other cases as well. When one writes fiction, is the intent to reference specific instances of possible entities with names such as "Harry Potter", "Thor", "Sherlock Holmes", and so on? It would seem not. The authors of fiction know full well that they are, precisely, writing fiction.

Information exhibited in our cases must be about something. But those who have sought to introduce "dummy" instances to support their information extraction proposals have offered mere promises of a workable solution. Some of their proposals do not generalize beyond specific cases; some force ontologists to make false assertions about relevant cases. Of course, all theories have their costs, and these costs may be worth bearing if no alternative solution is available that generalizes across our different cases and does not require insupportable ontological commitments. In fact, however, there is such a solution, one that leverages subtleties of OWL2DL; this solution will occupy us in the remainder of this article.

## 4. Unreal Patterns

Our central idea is that expressions putatively about non-existent entities should be modeled as **Information Content Entities** that are ultimately *about* – not non-existent dummy instances – but rather *logical combinations of actual classes*. Logical combinations within scope include the intersections, unions, and negations of classes, as described in the logical vocabulary for OWL2DL; what makes a class "actual" in the intended sense is that it is or has been instantiated.

The strategy defended here is inspired by the strategy sketched in [2] to explain how fictional expressions such as "Sherlock Holmes is a cocaine user" can inherit aboutness from components referenced in the expression, such as the string "cocaine" referring to the actual class **Portion of Cocaine**. It is, moreover, based on the strategy introduced in [5], applied to fictional entities in [39], and discussed in the context of digital twins in [42]. Our proposal differs from previous discussions in several respects. First, we generalize the strategy beyond potentially incorrect health care records [5] and fictional entities [39] to simulations and blueprints (and from there to plans, requirements specifications, government bills, historical documents, and many more). Second, we provide a recursive recipe to decompose representations of relevant scenarios to representations of actual classes – and we do this in more detail than in [2, 42]. Third, rather than adopt a sub-property of *is about* that signifies fictional entities: *as-if about* [39], we leverage resources from the Common Core Ontologies (CCO) [44] reflecting relations of *describing*, *prescribing*, and *representing* that allow us to distinguish our scenarios ontologically. Fourth, we expand the strategy to cover object properties reflecting relations that may have no real relata, such as *fires eye laser.* Fifth, we propose a solution to a puzzle identified in [39] regarding fictional entities and *a fortiori* to other cases where logical constraints on ontologies are violated.

*4.1 Specializations of Information*

To illustrate our proposal, we describe a recursive recipe for decomposing **Information Content Entities** that are not obviously about anything that exist into logical combinations of actual classes. Many of the ingredients have been described already, but we introduce the remaining ingredients here.

CCO provides three sub-properties of *is about* designed to reflect different attitudes agents may bear in regard to the relations holding between **Information Content Entities** and **Entities** which they are about. These object properties provide a foundation on which to represent paradigmatic examples of our cases, without requiring reference to problematic instances. These sub-properties[4] of *is about* are:

- x *describes* y iff x is an **Information Content Entity**, and y is an **Entity**, such that x *is about* the characteristics by which y can be recognized or visualized
- x *prescribes* y iff x is an **Information Content Entity** and y is an **entity**, such that x serves as a rule or guide for y if y is an **Occurrent**, or x serves as a model for y if y is a **Continuant**
- x *represents* y iff x is an instance of **Information Content Entity**, y is an instance of **Entity**, and z *is a carrier of* x, such that x *is about* y in virtue of there existing an isomorphism between characteristics of z and y

The information content of a newspaper article *describes* some current event, just as an accident report *describes* some accident. A blueprint *serves as a model for* some product, just as a professional code of conduct *serves as a set of rules for* anyone acting in the corresponding professional role. The content of a photograph *represents* the photographed entity, just as the content of a transcript *represents* the verbal interaction transcribed. The sense of "isomorphism" in the definition of *represents* is relative to the type of entities involved. For example, the arrangement of Napoleon's body parts in a painting by Jacques Louis David was meant to reflect the actual arrangement of these parts in Napoleon's body.

With respect to our cases, we maintain that fictional **Information Content Entities** are best understood as *describing* some logical arrangement of classes and object properties, ultimately in terms of actual classes. Blueprints, on the other hand, are best understood as **Information Content Entities** *prescribing* some arrangement of classes and object properties; and simulations are best understood as *representing*, insofar as, were the simulated phenomena to exist, so too would a relevant isomorphism.

[4] If $R_2$ is OWL2DL sub-property of $R_1$ then any pair <x, y> that is a member of the set $R_2$ is also a member of the set $R_1$. In other words, if x *describes* y then x *is about* y.

*4.2 Recursive Recipe*

With these ingredients in hand, we now turn to our recipe. To provide an ontological representation of an entity that did not, does not and perhaps will never exist:

1. Introduce an **Information Content Entity** for the entity, which is not about any actual instance modulo mental representations.
2. Identify classes and object properties which reflect the intended meaning of the **Information Content Entity** under 1, such as **Super Strength**, **Portion of Metal**, **Ground Vehicle**, **Kryptonian**, *inheres in*, *continuant part of*, *fires*, and so on.
3. Leverage OWL2DL to assert that the identified **Information Content Entity** *describes*/*prescribes*/*represents only* a class *C* that is equivalent to the logical combination of classes and restrictions on object properties articulated in 2, where canonical cases of:
    a. Fictions are said to *describe*,
    b. Blueprints are said to *prescribe*,
    c. Simulations are said to *represent*.
4. For any class or relata of any object property constituting *C* that has no instance, return to 2 and repeat.
5. Otherwise, when each class and relata of each object property constituting *C* or class and relata of each object property decomposed from those constituting C has at least one instance, stop.

Regarding 3: In OWL2DL, there are only three viable options for relating the relevant **Information Content Entity** to what it *is about*, by asserting: universal constraints (all x *is about* only y), existential constraints (all x *is about* some y), or a direct instance-to-instance (specific x *is about* specific y) relation. We can put aside the last, since it would require introducing a "dummy" instance which the relevant **Information Content Entity** *would be about*. We can also put aside the existential constraint, which amounts to asserting the relevant **Information Content Entity** *is about* some, though no particular, instance. This again runs counter to the intuition that in fiction, blueprints, and simulation relevant **Information Content Entities** are not necessarily about any given instance, since there often is no such instance. This leaves us with the universal constraint, which amounts to asserting that the relevant **Information Content Entity** *is about* only an instance falling within a class.

Importantly, this option neither asserts nor implies that there is such an instance. It requires only that, *if there were such an instance*, then it would have to be an instance of a class resulting from the logical combination of actual classes and object properties. It is this combination of classes that a relevant instance of **Information Content Entity** will be *about*. It is, moreover, *only* this combination that expressions putatively about such an instance will in fact be *about*.[5] One might, at this point, balk, given that we seem to have simply replaced "dummy instances" with "dummy classes". Not quite. What we have done is avoid introducing expressions putatively referring to instances but which in fact do not denote anything, by appealing to logical combinations of classes that themselves consist of instances that exist in the real world.

[5] We are also not introducing fictitious entities such as Superman as the result of logical combination of classes and object properties. Rather, we are asserting the knowledge contained in corresponding text materials can be characterized in terms of such combinations.

In every case, classes and object properties must be unpacked into actual classes and object properties that have only actual classes as relata.[6] Object properties are, in this recipe, ultimately explicated in terms of the classes which are their domains and ranges. We now apply this recipe for each of our paradigmatic examples.

*4.3 Superman*

Because we are dealing with fiction, we leverage the *describes* relation in our formalization. For simplicity, we introduce a subclass of **Information Content Entity (ICE)** called **Fictional Description**, as well as various classes and object properties where needed, at least one of which has no obvious parallel in the actual world. We then have, assuming a standard list of features associated with the character Superman:

- Superman description instance of **Fictional Description** and *describes* only
    - **Person** and
    - *described by* some **Superman Comic** and
    - *bearer of* some **Super Strength** and
    - *located in* value Earth and
    - *bearer of* some **Flight Disposition** and
    - *has origin* value Krypton and
    - *fires eye lasers* some **Laser** and …
- x *fires eye laser* y iff x instance of **Person** and *has part* some (**Eye** and *bearer of* **Laser Firing Disposition**) and y instance of **Laser** and …
- Krypton description instance of **Fictional Description** and *describes* only
    - **Astronomical Entity** and
    - *bearer of* some **Rocky Quality** and …

At the completion of our recursive application of the recipe, we end with "Superman" being defined ultimately in terms of classes and object properties that have actual instances. For each string putatively referring to some non-actual entity, such as "Krypton", or to some non-actual relation, such as "fires eye lasers", we decompose to classes all of which have instances, such as **Rocky Quality** or **Laser**.

Even so, one might wonder what to do if, rather than *fires eye lasers*, the pertinent object property would be *fires laser* having a class **Eye Laser** as its range, a class which has no actual instances. In this case, we would simply leverage the classes **Eye** and **Laser** to define the class **Eye Laser**, which would of course have no instances but would be decomposable into actual classes. This approach can be applied equally to object properties such as *fires eye laser* with range **Eye Laser**.

It has been suggested in [37] that such a recipe falters when fictional characters are represented as engaging with entities that violate constraints of imported ontologies. Following the example of [39], Superman might encounter a ghost which – one might argue – should be classified as a **Person** having material parts and yet also as an **Immaterial Entity** having no material parts. In BFO, the class **Material Entity** is disjoint with the class **Immaterial Entity**, so such a classification would result on our approach in an inconsistency. As a flat-footed response, note that while authors might label ghosts as persons, that does not mean ontologists should follow this labeling. Ontology engineers must not let themselves be tricked by mere labels. This is to say that, if there are examples of ghosts classified as persons, then it is plausible that the authors do not mean by "person" an entity with material parts. For such cases, we might introduce a class expression such as "ghost person", situated outside the material entity hierarchy.

[6] Note in most cases the recipe will result in necessary, but rarely sufficient, conditions.

This will only take us so far, of course. Authors of fiction need not abide by ontology best practices. We might envision an author who creates a fictional character that is explicitly a ghost that has no material parts and yet also, simultaneously, has material parts. This is a logical contradiction. Even then, however, there need be no problem for our approach. The relevant hierarchy could be extended to accommodate, perhaps with a subclass of **Continuant** each instance of which *has continuant part* some **Material Entity** and *has continuant part* some **Immaterial Entity**. Our hypothesized ghost description would then be about only instances of the **Continuant** subclass so defined, though this class would remain empty. For such a case, the above recipe will not terminate, which is also as it should be. There are descriptions of logical inconsistencies in the world, but they are not about any instances in the world.

*4.4 Honda Civic SLS 2035*

In explicating the blueprint case, we leverage the *prescribes* relation in our formalization and introduce a subclass of **Information Content Entity** called **Blueprint**.

- Honda civic SLS 2035 blueprint instance of **Blueprint** and *prescribes* only
    - **Ground Vehicle** and
    - *has continuant part* **Engine** and
    - *has continuant part* **Metal Chassis** and
    - *has continuant part* **Seat** and
    - *bearer of* some **Transportation Disposition** and
    - *has origin* value Tokyo Honda Factory and …

Moreover, any reference within **Blueprints** to other **Blueprints** can be characterized by following our decomposition recipe as illustrated with fiction above.

There is, however, a remaining question over the relationship such a pattern bears to prescribed entities once they exist. Applying the proposal for digital twins as defined here [42], we say that the **Blueprint** initially only *prescribes* that instances be created, though as yet no specific instance of the relevant class exists. Once an instance is created on the basis of the **Blueprint**, we say that this **Blueprint** still *prescribes* new instances to be created but also *represents* the instance that has already been created. The axioms governing *represents* in CCO would then entail that the **Blueprint** in question is a special type of **Information Content Entity**, namely, a **Representational Information Content Entity**. This class is not disjoint from its sibling – the domain of *prescribes* – which is **Directive Information Content Entity**. In other words, CCO allows for **Information Content Entities** to both *prescribe* and *represent* simultaneously.

*4.5 War Games*

In explicating simulation, we leverage the *represents* relation in our formalization, as well as a new type of **Information Content Entity** called **Cyber War Game Simulation Representation**.

- Red team simulation representation instance of **Cyber War Game Simulation Representation** and *represents* only
    - **Cyber-Attack Process** and
    - *has participant* value US Army and
    - *has participant* **Adversary** and
    - *has occurrent part* (**Act of Targeting** and has value Army Network 1 and *occupies temporal region* value March 23 2025) and
    - *has occurrent part* **Strategic Response Process** and (*has participant* value US Army Cyber Response Team) and (*occupies temporal region* value March 24, 2025) and …

One might worry that such hypothetical scenarios could represent contradictory events or impossible strategic responses—for instance, deploying centralized and decentralized defensive tactics simultaneously. Such hypothetical contradictions present no inherent difficulty provided that our representation refrains from enforcing unnecessary mutual exclusivity among these scenario elements, following suggestions similar to those discussed above with respect to fiction.

## 5. Excursus on the Future

Considering the future as an object of information raises important questions distinct from those addressed above. Statements about the future, such as weather forecasts or financial predictions, differ fundamentally in that they typically purport to describe events that are expected to materialize, rather than prescribing actions or representing hypothetical scenarios. Yet here, too, we are dealing with statements that are fundamentally dependent on context.

Unlike fictional entities, where it is clear no reference to actual instances is made, predictions about the future frequently concern events we expect to become actual [43]. Unlike blueprints, predictions about the future are not often taken to prescribe how we want the future to be. Future predictions are perhaps closer to simulations in this respect, as they may involve descriptions of what might happen were certain conditions to obtain. That said, temporal expressions appear to implicitly involve reference to temporal cycles, such as circadian rhythms, eating schedules, calendars, timelines on Gantt Charts, and so on.

Consider that the assertion “The Sun will rise tomorrow” implicitly refers to a certain temporal cycle observed in the actual world, known historically to recur, as evidenced by the regular succession of days and nights. It is indeed the same cycle that will be referred to tomorrow when I utter “The Sun will rise tomorrow”. This point deserves elaboration. Temporal speech often involves *indexicals* – expressions whose meaning varies from one context of use to another [45] – such as “tomorrow”, “next week”, “Friday”, “now”, and so on. I might write a Post-it note to hang on my door that reads “out of office Friday” and use it every week. Ontologically, I am using the same token inscription on the Post-it note and material entity bearer, but the note plausibly carries different meanings on Friday 7th, 2025 then Friday 14th, 2025, and so on.

Throughout, there is a cycle implicitly referenced in the use of "Friday", an expression that is about some instance of **Temporal Interval**, though which instance changes based on when and how the expression is used. For example, the **Information Content Entity** reflected by "out of office Friday" picks out a different instance of **Temporal Interval** each week. This is seen most clearly by observing that the instance of **Temporal Interval** picked out by "next Friday" differs from the instance of **Temporal Interval** picked out by "this Friday" insofar as the former stands in a *preceded by* relationship to the latter, but not vice versa. To accommodate these observations, we can here leverage from CCO another sub-property of *is about* distinct from those introduced above, namely:

- x *designates* y iff x is an **Information Content Entity**, and y is an **Entity**, such that given some context, x uniquely distinguishes y from other **Entities**

This need not force reference to a **Temporal Interval** that does not yet exist. We can, rather, rely on the recipe strategy outlined above to ensure that expressions about the future are, ultimately, ontologically unpacked in terms of logical combinations of actual classes. For example, introducing for simplicity a **Temporal Expression** subclass of **Information Content Entity** and a subclass of **Time Interval** for **Friday**, the expression "next Friday" spoken on June 6th, 2025, may be characterized as:[7]

- Friday expression instance of **Temporal Expression** and *designates* only
    - **Friday** and
    - *expressed on* value 2025-06-06T00:00:00 and
    - *preceded by* (**Temporal Instant** and *has date time* value 2025-05-06T00:00:00) and
    - *has first instant* (**Temporal Instant** and *has date time* value 2025-13-07T00:00:00) and …

Where the designated class has no instance on June 6th, 2025, but will have such an instance on June 13th, 2025. Strictly speaking, to satisfy the requirement for context and to uniquely distinguish a temporal expression from other **Entities**, as required by the definition of *designates*, one would need to add further clauses regarding precedence and context of utterance, but this is in principle achievable. The recipe thus carries over to expressions like "out of office next Friday" as described in the Post-it note example.

## 6. Conclusion

We have proposed a practical method for ontologically formalizing information content entities that putatively reference entities which do not exist or may never come to exist. Paradigmatic cases were: fictional entities, blueprints, simulations, and predictions about the future. Each case was explicated by employing specific sub-properties of aboutness, drawn from CCO: *describes*, *prescribes*, *represents*, and *designates*. Our presentation here should not suggest these are the only such cases within scope; our recipe applies equally to postulated entities in the natural sciences – such as the Higgs boson – at a time when it is not known whether the relevant entities exist, for example, by defining a combination class **Fundamental Particle** that is **Scalar**. Whether this class has instances remained, for a time, unknown. Yet we might argue that there was a time when crucial theoretical work in physics turned precisely on the formulation of hypotheses in its terms.

[7] The utterance "next Friday" would of course be a different utterance (a species of **process**) each week.

In this work, we understand information that is putatively not about anything that did, does or perhaps will exist, as **Information Content Entities** that are about logical combinations of actual classes and object properties, as opposed to hypothetical or non-existent instances. This strategy avoids ontological inconsistencies that arise from competing proposals, such as the introduction of "dummy instances", and appeals to modal contexts. Moreover, the proposal readily extends across diverse cases. In addressing potential challenges, such as contradictory blueprints or hypothetical scenarios, we maintain that such issues primarily involve labeling rather than structural inconsistencies. By introducing carefully crafted subclasses, it is straightforward to maintain consistency in ontology representations. We intend our unified approach to contribute to resolving longstanding complexities associated with modeling such entities within formal ontologies and look forward to feedback from the broader community on this and related topics.